\documentclass[letterpaper, 10 pt, conference]{IEEEtran}  
\usepackage{amsmath, amssymb, graphicx, caption, subcaption, float, kotex, booktabs, array, svg, pdfpages, xcolor, color, colortbl, multirow, balance, makecell, caption}
\usepackage{verbatim}
\usepackage{rotating}
\usepackage{pifont}
\usepackage[normalem]{ulem}

\usepackage{amsmath}
\usepackage{amssymb}

\DeclareFontFamily{U}{stix2bb}{}
\DeclareFontShape{U}{stix2bb}{m}{n} {<-> stix2-mathbb}{}

\NewDocumentCommand{\indicator}{}{\text{\usefont{U}{stix2bb}{m}{n}1}}

\IEEEoverridecommandlockouts                              
\definecolor{LightCyan}{rgb}{0.88,1,1}

\title{\LARGE \bf
SyNeT: Synthetic Negatives for Traversability Learning}

\author{Bomena Kim, Hojun Lee, Younsoo Park, Yaoyu Hu, Sebastian Scherer and Inwook Shim$^\dagger$}

\begin{document}

\maketitle
\thispagestyle{empty}
\pagestyle{empty}

\begin{abstract}

Reliable traversability estimation is crucial for autonomous robots to navigate complex outdoor environments safely. Existing self-supervised learning frameworks primarily rely on positive and unlabeled data; however, the lack of explicit negative data remains a critical limitation, hindering the model's ability to accurately identify diverse non-traversable regions. To address this issue, we introduce a method to explicitly construct synthetic negatives, representing plausible but non-traversable, and integrate them into vision-based traversability learning. Our approach is formulated as a training strategy that can be seamlessly integrated into both Positive–Unlabeled~(PU) and Positive–Negative~(PN) frameworks without modifying inference architectures. Complementing standard pixel-wise metrics, we introduce an object-centric FPR evaluation approach that analyzes predictions in regions where synthetic negatives are inserted. This evaluation provides an indirect measure of the model's ability to consistently identify non-traversable regions without additional manual labeling. Extensive experiments on both public and self-collected datasets demonstrate that our approach significantly enhances robustness and generalization across diverse environments. The source code and demonstration videos will be publicly available.

\end{abstract}
\section{INTRODUCTION}
With advances in visual perception, traversability estimation in outdoor environments has become a key component of autonomous mobile robots. Traversability refers to the robot’s ability to determine, based on visual inputs, whether a region is safe to traverse. While many studies have explored both traditional rule-based methods~\cite{wermelinger2016navigation, fankhauser2014robot} and supervised learning approaches~\cite{pan2024traverse, ahtiainen2017normal, Sharma2022CaT}, applying them to real-world scenarios remains difficult. The main challenge lies in the complexity and variability of real-world outdoor driving conditions, which make acquiring a comprehensive labeled dataset prohibitively difficult. Since supervised methods require precise pixel-level annotations for training, the high cost and labor of data acquisition become a major bottleneck, limiting the scalability of these approaches.

To alleviate the heavy reliance on large-scale manual annotations, self-supervised learning~(SSL) methods~\cite{zurn2020self, cho2024learning, chen2023learning, jeon2024follow, wellhausen2019should, seo2023scate, schmid2022self, cai2025pietra} have emerged as a promising alternative. In these approaches, interaction signals along the robot’s driving trajectories, such as footprints or wheel tracks, are extracted and utilized as robot-derived supervision for identifying traversable surfaces. These interaction samples are treated as positive~(traversable), while the remaining, unexplored regions are kept unlabeled. By leveraging strategies like pseudo-labeling to compensate for the absence of explicit negative labels, these SSL methods have demonstrated robust performance in outdoor scenarios.

However, these methods still struggle to identify non-traversable regions. First, pseudo-labeling on unlabeled data does not resolve the inherent uncertainty of non-traversability. As a result, SSL methods continue to produce blurry boundaries between traversable and non-traversable regions and struggle to predict unsafe areas. Second, naturally collected driving experiences rarely contain sufficient failure cases, making it difficult to obtain true negative samples. This lack of negative data biases the learning process, ultimately hindering robust adaptation to diverse driving situations, including both off-road and complex social environments.

\begin{figure}
    \centering
    \captionsetup{font=small}
    \includegraphics[width=0.48\textwidth]{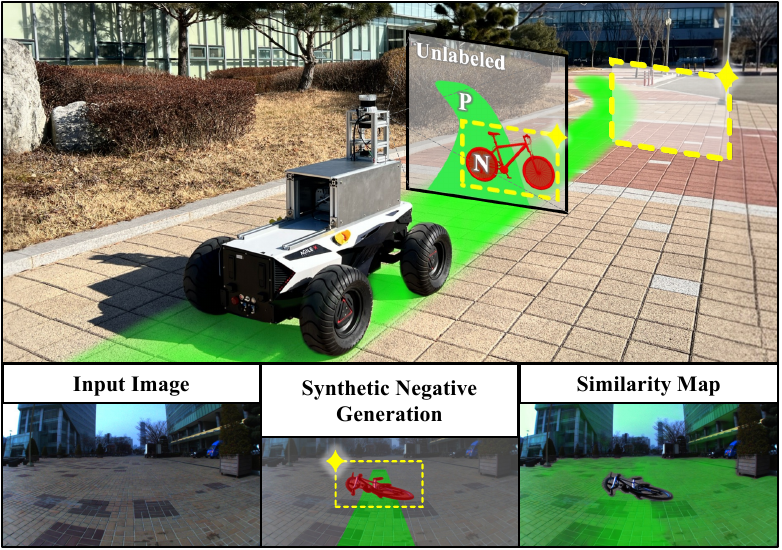}
    \caption{High-level concept of our synthetic negative-based traversability estimation. 
    (Top) The robot treats its driving path as positive and introduces synthetic objects as explicit negatives. (Bottom) The pipeline takes an input image, generates a synthetic negative, and produces a similarity map for traversable regions.}
    \label{fig_hunter}
\end{figure}

\begin{figure*}[t]
    \centering
    \captionsetup{font=small}
    \includegraphics[width=\textwidth]{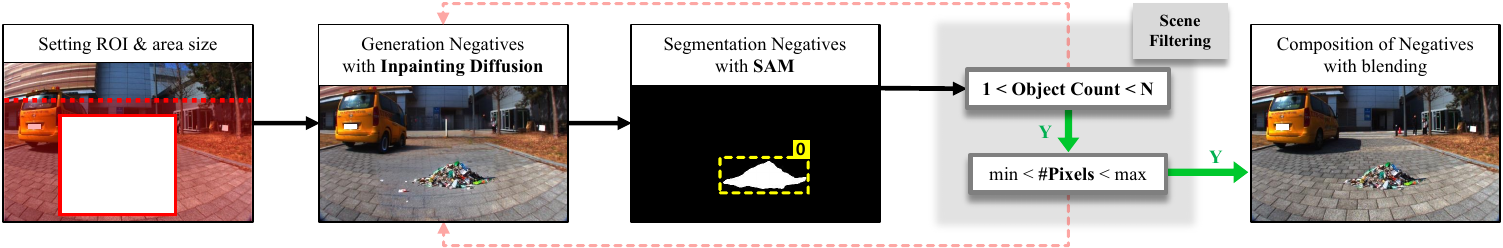}
    \caption{
    Synthetic negative generation pipeline. (1)~Region selection: randomly sample a region of interest~(ROI) and target object size within the predefined ground area. (2)~Inpainting: generate synthetic negatives using the Stable Diffusion 3.5~\cite{stabilityai_sd35_2024} inpainting pipeline and FLUX.1 Fill~\cite{bfl_flux1_tools_2024}. (3)~Segmentation \& filtering: segment the generated object with LangSAM~\cite{medeiros_lang_segment_anything_2023} and apply scene filters~(\textit{e.g.}, object count and pixel-area thresholds); if unmet, repeat steps (2) and (3) for the same ROI. (4)~Composition: blend the approved negative object into the base image to obtain the final composite and its negative mask.
    }
    \vskip -0.2em
    \label{fig_diffusion}
\end{figure*}

Recent studies~\cite{jung2024v, ma2025imost} have shown that visual foundation models~(VFMs) can help reduce the uncertainty of unlabeled regions in self-supervised traversability learning. These methods select initial traversable seeds from the robot’s past driving trajectories and use a VFM, such as Segment Anything~(SAM)~\cite{kirillov2023segment}, to expand those seeds to visually similar regions. Thus, some of the previously unlabeled regions are treated as traversable, while the remaining regions are used as negative cues. However, these regions are not explicitly verified as non-traversable and often contain ambiguous or even traversable pixels, producing false-negative supervision. Consequently, the resulting negative labels remain low-confidence and do not fully resolve non-traversable supervision.

To further address the lack of non-traversable regions, other studies~\cite{bae2023self, bu2025self, schreiber2024w, schreiber2025you} pursue negative supervision to help distinguish hazardous terrain. Some construct manually labeled non-traversable datasets~\cite{bae2023self, schreiber2024w, schreiber2025you}, while another uses LiDAR to automatically detect obstacles and mark them as negative regions~\cite{bu2025self}. However, reliance on manual annotation or rule-based detection limits scalability and adaptability in complex, diverse environments.

In this paper, we propose SyNeT: Synthetic Negatives for Traversability Learning, which is a training strategy designed to address the scarcity of explicit negatives in self-supervised traversability learning. Unlike conventional methods, our approach explicitly constructs synthetic negatives and integrates them into the training objective, enabling the model to learn robust traversability features without architectural modifications. Furthermore, we leverage the generated synthetic negatives to establish an automatic evaluation approach. This method measures the object-centric False Positive Rate (FPR) on inserted negatives, providing a quantitative measure of the model’s ability to correctly identify non-traversable regions without additional labeling costs. Finally, extensive experiments on public benchmarks and a self-collected dataset with manual annotations demonstrate that our approach significantly enhances robustness and generalization across diverse environments, including off-road, urban and social settings. In particular, by reducing false positives on non-traversable regions, our approach can support safer robot navigation in complex environments. To provide reproducibility and field deployment, all training/inference code and TensorRT-optimized runtimes will be released.

\begin{figure*}[t]
    \centering
    \captionsetup{font=small}
    \includegraphics[width=\textwidth]{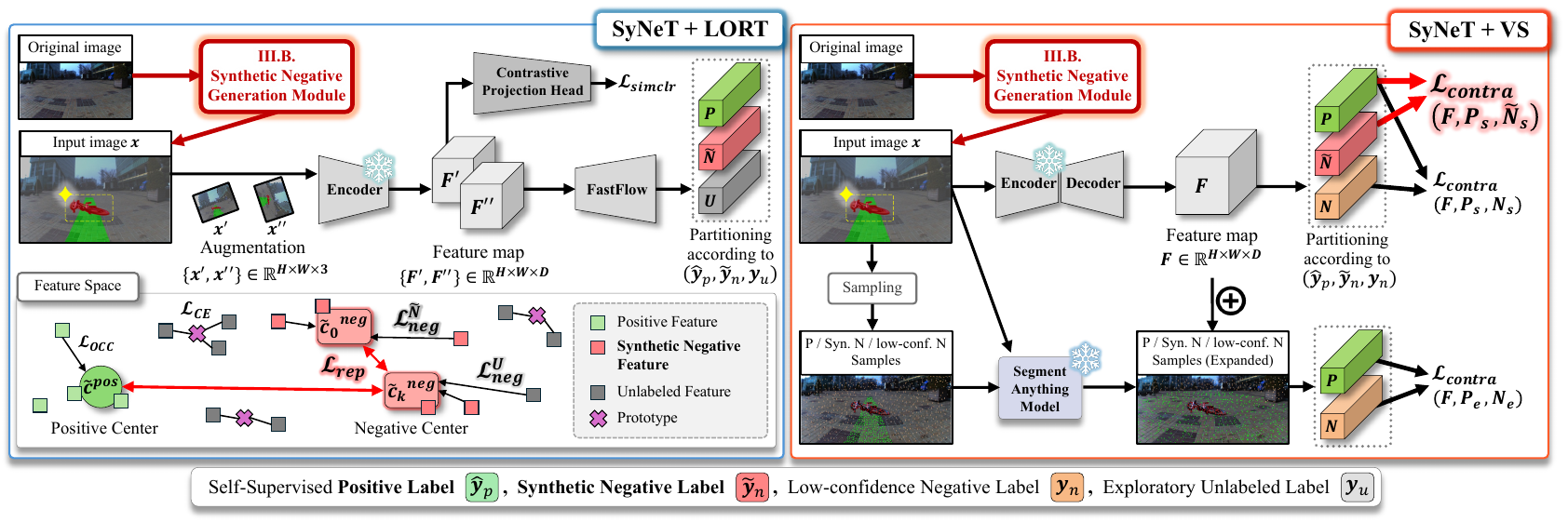}
    \caption{Overview of the SyNeT training strategy integrated into LORT~(PU) and VS~(PN) baselines. The Synthetic Negative Generation Module augments the input image with explicit negatives, allowing the model to extract synthetic negative features~($\widetilde{N}$). These features are then incorporated into the respective learning objectives~($\mathcal{L}_\mathrm{neg}$ for LORT, $\mathcal{L}_\mathrm{contra}$ for VS) to enforce a clear separation between traversable and non-traversable regions in the feature space.}
    \label{fig:overview}
\end{figure*}

\section{Related Works}

\subsection{Positive–Unlabeled Learning Approaches}
Recent work on traversability estimation has increasingly relied on self-supervised learning to reduce the need for manual annotations. Most approaches follow either Positive–Unlabeled~(PU) learning~\cite{bekker2020learning} or one-class learning~\cite{ruff2018deep}. These methods use only positive and unlabeled data and avoid explicit labeling of non-traversable regions. LORT~\cite{seo2023learning} is a representative PU-based method that clusters unlabeled regions and regularizes PSPNet features~\cite{zhao2017pyramid} toward these clusters. This helps reduce bias toward the positive class and prevents feature collapse. More recently, VFMs have been explored for traversability estimation~\cite{aegidius2025watch, kim2024learning}. STEPP~\cite{aegidius2025watch} extracts dense features from a pretrained VFM (DINOv2~\cite{oquab2023dinov2}). It then trains an MLP-based encoder-decoder on these features and uses reconstruction error as a traversability cue. Together, these approaches mitigate label scarcity without explicit negatives, but they still provide only indirect modeling of non-traversable regions, leaving decision boundaries sensitive to data bias and rare obstacles.

\subsection{Positive–Negative Learning Approaches}

In contrast to PU learning, Positive–Negative~(PN) approaches incorporate negative information for non-traversable regions into traversability learning~\cite{jung2024v, bae2023self, bu2025self, schreiber2024w, schreiber2025you}. 
These methods use both positive and negative supervision for training, rather than relying on positive and unlabeled data. 

V-STRONG~\cite{jung2024v} is a self-supervised PN method that derives implicit negative information from the complement of trajectory-based traversable regions. The method first selects traversable regions from past driving trajectories and expands them to visually similar regions using a pretrained VFM~(SAM~\cite{kirillov2023segment}). The complement of the SAM-expanded traversable regions is used as a source of negative candidates. However, these regions are not explicitly verified as non-traversable and can still include ambiguous or even traversable pixels. This ambiguity undermines the reliability of the negative supervision for non-traversable regions.

CHUNGUS~\cite{schreiber2025you} is a human-in-the-loop traversability learning framework that incorporates human supervision during training. It relies on sparse human-provided ordinal comparisons to guide the model, without explicitly specifying non-traversable regions. Since these comparisons only indicate which regions are more traversable, negative information is provided implicitly. Additionally, human input is required whenever the robot encounters new environments, leading to substantial human labeling costs.


Bu \textit{et al}.~\cite{bu2025self} propose a self-supervised traversability method that derives non-traversable regions in bird’s-eye view~(BEV) using LiDAR-based obstacle detection. These regions are labeled as non-traversable and used as explicit negatives for training. Compared to prior methods that rely on implicit negative supervision~\cite{jung2024v,schreiber2024w, schreiber2025you}, this approach leverages geometric cues from LiDAR to construct explicit non-traversable labels. However, the quality of the negative supervision depends on the accuracy of the obstacle detection and is limited to obstacles observed during data collection.

In contrast, our work explicitly constructs synthetic negatives directly in the image space and integrates them into the training objective, enabling robust traversability learning without additional sensors or architectural modifications.

\section{Methods}

SyNeT consists of three components: (1)\,positive labels derived from executed trajectories, (2)\,synthetic negatives generated with a diffusion model, and (3)\,learning pipelines that integrate both positive and synthetic negative labels. First, reliable traversability cues are obtained without manual annotation by leveraging the robot’s driving experience~(Sec.~\ref{sec_label}). Next, a diffusion-based negative data generation module inserts semantically coherent negatives into real scenes to provide explicit negative supervision~(Sec.~\ref{sec_diffusion}). Finally, the training strategy jointly utilizes real and synthetic data, with losses designed for stable, consistent estimation~(Sec.~\ref{sec_learning}).

\subsection{Self-Supervised Traversability Label}
\label{sec_label}
Following prior studies~\cite{jung2024v,bu2025self,seo2023learning}, the self-supervised traversability label~$\hat{\mathbf{y}}_p$ is obtained from images recorded during robot operation. Regions actually traversed by the robot are labeled positive, using ground-contact points obtained from SLAM-reconstructed trajectories. The contact representation follows the locomotion type~(\textit{e.g.}, wheel tracks for wheeled platforms and footprint for legged robots).

\subsection{Synthetic Negative Data Generation}
\label{sec_diffusion}
We propose a diffusion-based pipeline to generate scene-consistent synthetic negatives for traversability learning. As shown in Fig.~\ref{fig_diffusion}, the pipeline has three stages: region selection, object generation, and composition. The module produces scene-coherent composites and pixel-wise negative masks. 

This pipeline emphasizes scene-level coherence rather than object structure. This is because, in a pixel-wise feature learning framework, this global consistency is more important for learning stable traversability cues than fine geometric fidelity of the inserted negatives~(see Ablation Study~\ref{ablation2}).

With these scene-consistent synthetic negatives, we can explicitly supervise the model to treat the inserted negatives as non-traversable during training. The synthetic negative $\mathbf{\widetilde{y}}_n$ is then incorporated into the learning objective so that it directly contributes to separating non-traversable regions. In addition, the synthetic negatives are incorporated into the original annotations to produce a synthetic ground truth, making them directly usable for evaluation. This yields two evaluation settings: (1) Original~(standard human-labeled test set) and (2) Synthetic~(the same images augmented with synthetic negatives and their synthetic ground truth). We adopt this protocol in our experiments; see Sec.~\ref{sec:exp} for details.

\subsection{Traversability Learning}
\label{sec_learning}
We instantiate SyNeT within two representative traversability frameworks, Positive–Unlabeled~(PU) and Positive–Negative~(PN), and detail their training objectives. Fig.~\ref{fig:overview} illustrates the high-level pipelines for the PU and PN instantiations, respectively. The learning objective integrates synthetic negatives~$\mathbf{\widetilde{y}}_n$ with self-supervised positive labels~$\hat{\mathbf{y}}_p$, low-confidence negative labels~$\mathbf{y}_n$, unlabeled labels~$\mathbf{y}_u$ to form a feature space where traversable and non-traversable regions are separated with stable margins. Given an input image $x \in \mathbb{R}^{H \times W \times 3}$, the backbone network extracts a feature map $F \in \mathbb{R}^{H \times W \times D}$, which is then partitioned into positive, synthetic-negative, low-confidence negative and unlabeled sets according to the provided labels. Specifically, under the PU setting, the feature map is divided as $F \in \{P, \widetilde{N}, U\}$, whereas under the PN setting it is divided as $F \in \{P, \widetilde{N}, N\}$. All feature vectors are $\ell_2$-normalized before computing similarities.

\noindent\textbf{Positive-Unlabeled with Synthetic Negative:}
LORT~\cite{seo2023learning}, the recent state-of-the-art PU approach, builds a global positive center and a set of unlabeled prototypes, and is trained with three loss terms:
\begin{equation}
\label{eq_pu}
\mathcal{L}_\mathrm{LORT}
=
(1-\tau)\mathcal{L}_\mathrm{OCC}
+
\tau\mathcal{L}_\mathrm{CE}
+
\mathcal{L}_\mathrm{simclr},
\end{equation}
where $\mathcal{L}_\mathrm{OCC}$ increases the reconstruction likelihood of positive features, $\mathcal{L}_\mathrm{CE}$ learns the soft assignment of unlabeled features via prototypes, and $\mathcal{L}_\mathrm{simclr}$ is a contrastive term that enforces feature consistency across augmentations.

To incorporate synthetic negative features~$\widetilde{N}$ and prior PU learning formulation~(Eq.~\ref{eq_pu}), we introduce a negative-center assignment loss $\mathcal{L}_{neg}$. This loss softly assigns synthetic negative~$\widetilde{N}$ and the unlabeled features~$U$ to negative centers, thereby establishing a stable non-traversable reference within the original positive–unlabeled embedding space. Let $\{c_k^{neg}\}_{k=1}^{K}$ denote $K$ negative centers and let $\tilde{f}_i=f_i /||f_i||_2$, ${\tilde{c}}_k^{neg}=c_k^{neg}/||c_k^{neg}||_2$. For a feature set $F\in \{P, \widetilde{N}, U\}$, the negative-center loss is
\begin{equation}
\mathcal{L}_{neg}^{(F)}
=
-\frac{1}{|F|}
\sum_{f_i \in F}
\sum_{k}^{K}
T_{i,k}\,
\log
\frac{
\exp(\langle \tilde{f}_i , {\tilde{c}}_k^{neg} \rangle)
}{
\sum_{j}^{K} \exp(\langle \tilde{f}_i , {\tilde{c}}_j^{neg} \rangle)
},
\end{equation}
where $\langle\cdot,\cdot\rangle$ cosine similarity, $f_i$ a feature vector at $i_{th}$ pixel, and $T_{i,k}\in[0,1]$ with $\sum_{k} T_{i,k}=1$ are soft assignment targets.

The combined negative loss is
\begin{equation}
\mathcal{L}_{neg}
=
\lambda_N\mathcal{L}_{neg}^{\widetilde{N}}
+
\lambda_U\mathcal{L}_{neg}^{U}
, (\lambda_N,\lambda_U \geq 0; \lambda_N+\lambda_U=1 ),
\end{equation}
where $\lambda_N=\lambda_U=0.5$ as a default.
For synthetic negatives $\widetilde{N}$, $T_{i,k}$ is obtained from a softmax-based responsibility over centers. For unlabeled features $U$, $T_{i,k}$ is computed via a Sinkhorn~\cite{cuturi2013sinkhorn} assignment on a cost matrix derived from similarities, encouraging balanced usage of negative centers.

In parallel, we also add $\mathcal{L}_{rep}$ to prevent the positive and negative centers from collapsing toward each other and also encourage dispersion among negative centers.
The repulsion loss consists of both negative–positive and negative–negative interactions and is defined as:
\begin{equation}
\begin{split}
    \mathcal{L}_{rep} = & -\frac{1}{K} \sum_{k} \log\!\left( 1 - \langle {\tilde{c}}_k^{neg} , {\tilde{c}}^{pos} \rangle \right) \\
    & -\gamma\frac{1}{K(K-1)} \sum_{i \neq j} \log\!\left( 1 - \langle {\tilde{c}}_i^{neg} , {\tilde{c}}_j^{neg} \rangle \right),
\end{split}
\end{equation}
where ${\tilde{c}}^{pos}$ is the normalized positive center, and $\gamma$ controls the relative weighting between the two repulsion terms.
The final objective is given by:
\begin{equation}
\mathcal{L}_\mathrm{Ours}
=
\mathcal{L}_\mathrm{LORT}
+
\lambda_{neg}\mathcal{L}_{neg}
+
\lambda_{rep}\mathcal{L}_{rep},
\end{equation}
with weight $\lambda_{neg}, \lambda_{rep} \in [0, 1]$.

\noindent\textbf{Positive-Negative with Synthetic Negative:}
We instantiate SyNeT on V-STRONG (VS)~\cite{jung2024v}, a recent state-of-the-art PN framework. VS performs contrastive learning by sampling pixel-level features from trajectory-based and SAM-based sample sets. Specifically, trajectory-based supervision yields positive and low-confidence negative samples $(P_{\mathrm{s}}, N_{\mathrm{s}})$, and SAM-derived expanded regions produce $(P_{\mathrm{e}}, N_{\mathrm{e}})$.
The loss consists of the following two terms:
\begin{equation}
\begin{split}
\mathcal{L}_{\mathrm{VS}}
=
& \hspace{4pt}(1-\omega_{\mathrm{e}})
\,\mathcal{L}_{\mathrm{contra}}(F, P_{\mathrm{s}}, N_{\mathrm{s}}) \\
& +\omega_{\mathrm{e}}
\,\mathcal{L}_{\mathrm{contra}}(F, P_{\mathrm{e}}, N_{\mathrm{e}}),
\end{split}
\end{equation}

\noindent{where each term corresponds to the trajectory-based and SAM-mask–based contrastive losses, and $\omega_{\mathrm{e}} \in [0,1]$ controls the relative weighting between the two.}

We incorporate synthetic negative features $\widetilde{N}$ to introduce explicit non-traversable cues that closely resemble real obstacles. By providing explicit negative supervision, our synthetic negatives serve as reliable negative anchors, complementing the implicit negatives used in the original PN formulation of VS.
The contrastive loss using synthetic negatives is defined as:
\begin{flalign}
&\quad\mathcal{L}_{\mathrm{contra}}(F, P, \widetilde{N})=&
\end{flalign}
$$
-\frac{1}{|P| (|P| - 1)}
\sum_{q_i \in P}
\sum_{q_j \in P}
\indicator (i \neq j)
\log
\frac{
\exp\!\big(F_{q_i}^\top F_{q_j} / \tau\big)
}{
\sum_{q_k \in \widetilde{N}}
\exp\!\big(F_{q_i}^\top F_{q_k} / \tau\big)
},
$$
where $|P|$ is the number of positive samples, $F_q$ is the feature vector at pixel location $q$, and $\tau$ is the temperature scalar.
The final objective is given by:

\begin{equation}
\mathcal{L}_\mathrm{Ours}
=
\mathcal{L}_\mathrm{VS}
+
\lambda_n\mathcal{L}_{\mathrm{contra}}(F, P_\mathrm{s}, \widetilde{N}_\mathrm{s}),
\end{equation}
with weight $\lambda_n \in [0, 1]$.

\section{Experiments}
\label{sec:exp}
To assess compatibility and effectiveness with existing self-supervised traversability approaches, we evaluate public benchmark datasets and a self-collected dataset across diverse conditions, and introduce an object-centric evaluation that measures False Positive Rate~(FPR) on negative regions without any additional human-labeling. This metric offers a finer-grained analysis complementing image-level global scores, which average over the entire image and consequently dilute errors on non-traversable areas.

\begin{figure}[t]
\centering

\scriptsize
\captionsetup{font=small}
\captionof{table}{Overview of the datasets used for training and quantitative evaluation in our experiments}
\label{tab:dataset}
\setlength{\tabcolsep}{3.0pt}

\begin{tabular}{l|c|c|c|c|c|@{}}
\toprule
Dataset & \#Frame & \#Train & \#Test Original & \#Test Synthetic & Environments \\ \midrule
SCAND     & $\sim930k$    & $0$      & $0$    & $0$    & Social   \\
CEAR      & $\sim158k$ & $7,055$  & $0$    & $0$    & Indoor   \\
RELLIS-3D & $13,556$ & $10,295$ & $900$  & $1800$ & Off-road \\ \midrule
RUOS      & $\sim194k$ & $9,392$  & $1,000$& $400$  & Urban    \\
\bottomrule
\end{tabular}

\vspace{0.6em} 

\captionsetup{font=small}
\includegraphics[width=0.49\textwidth]{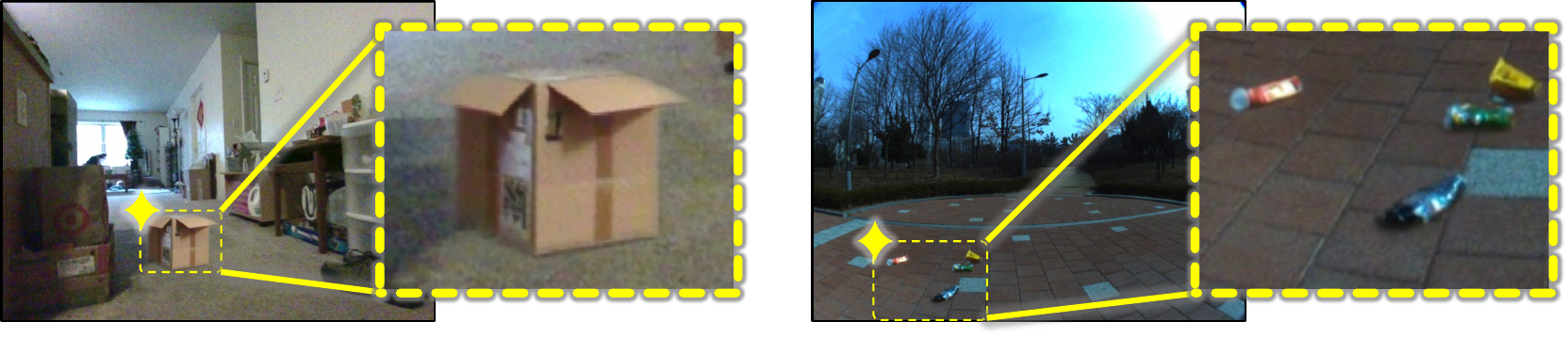}
\captionof{figure}{Example synthetic negatives inserted into real scenes.}
\label{fig:synthetic_examples}

\end{figure}
\begin{figure}[t]
    \centering
    \captionsetup{font=small}
    \includegraphics[width=0.48\textwidth]{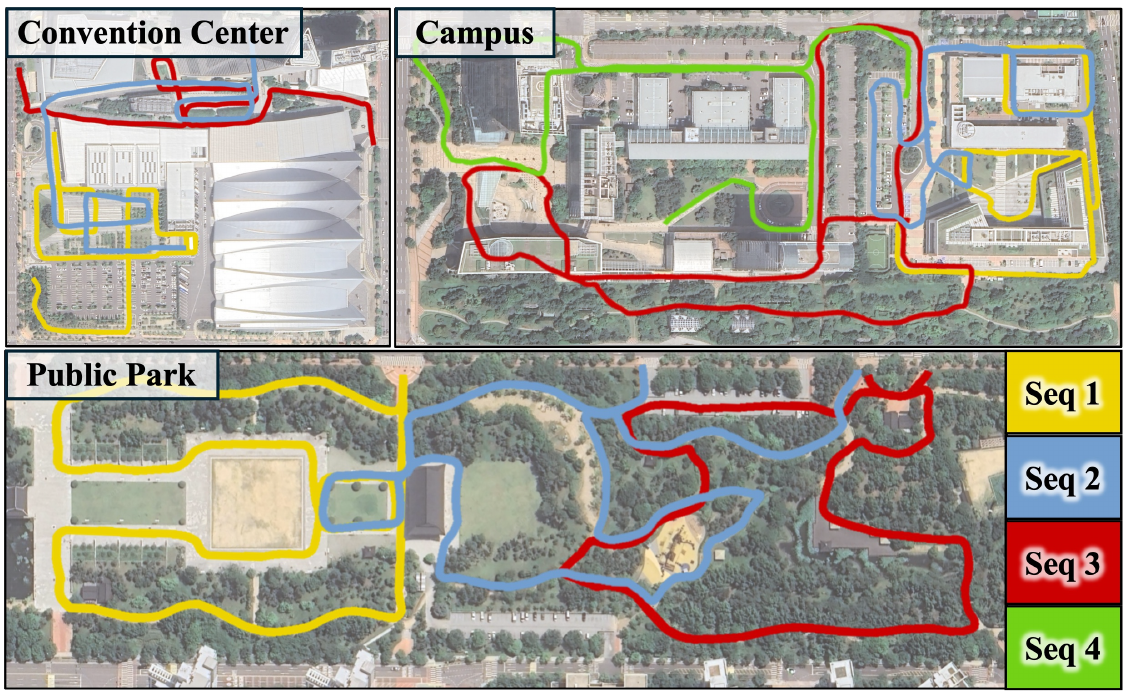}
    \caption{Overview of the RUOS dataset. 
    The driving trajectories are visualized across three distinct scenarios: Campus, Convention Center, and Public Park. Each color corresponds to a recording sequence, highlighting the diversity of the collected paths.}
    \label{fig:mrinha}
\end{figure}
\subsection{Datasets}
We evaluate our method on four representative datasets under both original and synthetic negative settings, as summarized in Table~\ref{tab:dataset}, with qualitative examples of synthetic negative generation shown in Fig.~\ref{fig:synthetic_examples}. To ensure a fair comparison, we fix the training set size and replace a fraction of the original images with synthetic negatives.  The columns \#Test Original and \#Test Synthetic denote, respectively, the number of human-labeled test images, and the number of test images augmented by inserting synthetic negatives~(Sec.~\ref{sec_diffusion}).

\noindent\textbf{SCAND.}~\cite{karnan2022socially}
A large-scale social navigation dataset with dense human presence. We use SCAND solely for qualitative analysis to examine traversability predictions in highly crowded social environments.

\noindent\textbf{CEAR.}~\cite{zhu2024cear}
A public indoor/outdoor dataset. We use only the indoor-home scenario and insert synthetic negatives, including tables, delivery boxes, small pets, and people, to emulate cluttered, socially dynamic settings; all human/animal instances are synthetically generated. Data is recorded with a quadruped robot.

\noindent\textbf{RELLIS-3D.}~\cite{jiang2021rellis}
A public off-road dataset with diverse terrain. We treat grass, dirt, asphalt, concrete, and mud as traversable. To evaluate generalization to non-traversable cases, we synthetically insert off-road obstacles such as rocks, logs and tree stumps, into each scene using our pipeline.

\noindent\textbf{RUOS~(Ours).}
Robots in Urban, Off-road and Social environments (RUOS) is a self-collected dataset for traversability learning in diverse real-world outdoor environments. It spans campus, convention center, and public park areas, covering urban, social, and off-road settings with diverse pavement textures. An overview of the data collection sites and trajectories is shown in Fig.~\ref{fig:mrinha}. We collected real-world data using a custom-built mobile robotic platform and inserted socially common synthetic negatives, including trash, boxes, bicycles, animals, and people, to examine robustness and sensitivity to negative regions.

\subsection{Implementation Details}

For consistent evaluation, we implement our SyNeT on two distinct baselines: LORT and VS. Across all configurations, we adopt a frozen image encoder strategy, optimizing the models using the Adam optimizer with an initial learning rate of $10^{-3}$. Standard data augmentations, including random cropping, flipping, and color jittering, are applied after resizing images to a fixed resolution. During training, synthetic negatives are injected into $20\%$ of the training samples, ensuring at least one synthetic instance per selected image. This ratio is kept consistent across methods.

\textit{LORT-based Configuration:} For the LORT-based implementation, we utilize both ResNet-50~\cite{he2016deep} and RADIO~\cite{heinrich2025radiov25improvedbaselinesagglomerative} as backbone feature extractors, with their weights kept frozen. Following the original LORT protocol, the model is trained for $60$ epochs with a batch size of $8$. Specific to SyNeT, we use four negative centers~($K=4$), set the loss temperature $\tau$ to $0.55$, and the repulsion weighting factor $\gamma$ to $0.1$.

\textit{VS-based Configuration:} For the VS-based implementation, we employ the ViT-H~\cite{dosovitskiy2020image} encoder from SAM~\cite{kirillov2023segment} as the frozen backbone. Adhering to the original VS training regime, this model is trained for $5$ epochs with a batch size of $1$. The negative loss weight $\lambda_n$ is set to $1.2$.

All experiments are conducted on a single NVIDIA RTX A6000 GPU. For reproducibility, the full codebase and trained models will be released at a public repository.

\begin{table}[t]
\centering
\scriptsize
\captionsetup{font=small}
\caption{Quantitative results on the RELLIS-3D.}
\setlength{\tabcolsep}{4.1pt}
\begin{tabular}{l|l|c|c|c|c|c|c|c|c@{}}
\toprule
 & Methods & AUROC$\uparrow$ & MaxF$\uparrow$  & AP$\uparrow$ & PRE$\uparrow$   & REC$\uparrow$   & FPR$\downarrow$   & FNR$\downarrow$   \\ \midrule
 & Schmid~\cite{schmid2022self} & 0.713 & 0.601 & 0.666 & 0.513 & 0.727 & 0.455 & 0.273 \\ 
 & LORT~\cite{seo2023learning} & 0.935 & 0.911 & 0.949 & 0.920 & 0.890 & 0.100 & 0.110 \\
 & VS~\cite{jung2024v} & 0.967 & 0.933 & 0.970 & 0.925 & 0.935 & 0.097 & 0.065 \\  \cmidrule(r){2-9}
 {\begin{rotate}{90}\footnotesize{Original}\end{rotate}} & \textbf{SyNeT}+LORT & \textbf{0.979} & \textbf{0.951} & \textbf{0.976} & \textbf{0.936} & \textbf{0.970} & \textbf{0.087} & \textbf{0.030} \\
 & \textbf{SyNeT}+VS & \textbf{0.973} & \textbf{0.945} & \textbf{0.980} & \textbf{0.938} & \textbf{0.951} & \textbf{0.083} & \textbf{0.049} \\ \midrule
 & LORT~\cite{seo2023learning} & 0.903 & 0.877 & 0.917 & 0.896 & 0.842 & 0.117 & 0.158 \\ 
 & VS~\cite{jung2024v} & 0.956 & 0.918 & 0.957 & 0.895 & 0.939 & 0.133 & 0.061 \\ \cmidrule(r){2-9} 
 & \textbf{SyNeT}+LORT & \textbf{0.966} & \textbf{0.923} & \textbf{0.966} & \textbf{0.914} & \textbf{0.909} & \textbf{0.103} & \textbf{0.091} \\
 {\begin{rotate}{90}\footnotesize{~Synthetic}\end{rotate}} & \textbf{SyNeT}+VS & \textbf{0.968} & \textbf{0.935} & \textbf{0.973} & \textbf{0.922} & \textbf{0.948} & \textbf{0.097} & \textbf{0.052} \\ \bottomrule
\end{tabular}%
\vspace{-3mm}
\label{tab:rellis_main}
\end{table}
\begin{figure*}[t]
    \centering
    \captionsetup{font=small}
    \includegraphics[width=0.85\textwidth]{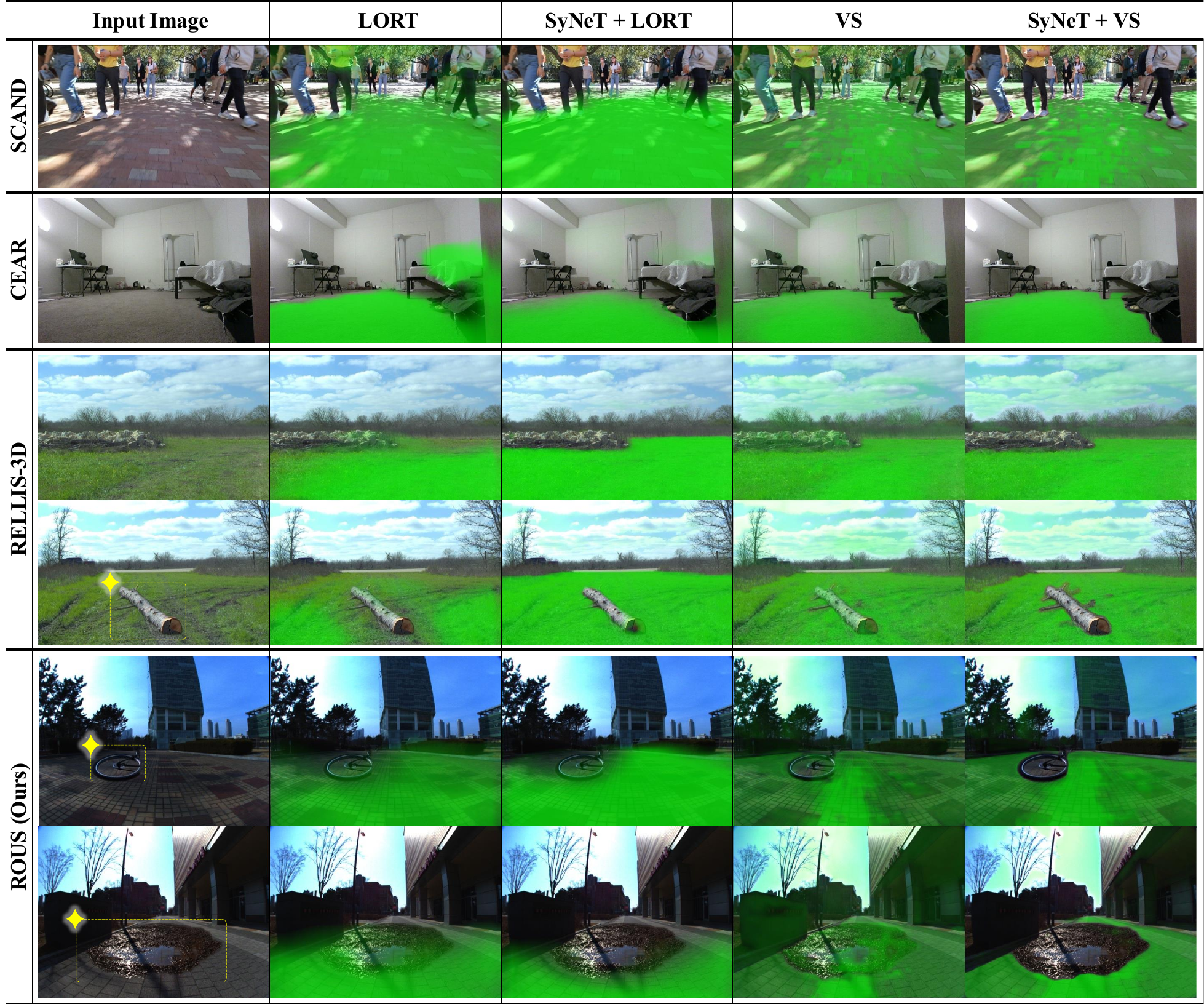}
    \caption{Qualitative comparison on SCAND, CEAR, RELLIS-3D and RUOS~(left→right: Input, LORT, SyNeT+LORT, VS, SyNeT+VS). }
    \label{fig:qualitative}
\end{figure*}
\subsection{Quantitative \& Qualitative Results}

Table~\ref{tab:rellis_main} reports performance under two evaluation settings on RELLIS-3D dataset: Original and Synthetic. The results demonstrate that both SyNeT+LORT and SyNeT+VS consistently improve their respective baselines across all metrics and scenarios. Especially, SyNeT substantially reduces both FPR and FNR in both settings, indicating a clearer separation between traversable and non-traversable regions. This improvements demonstrates that synthetic negative supervision is complementary to both learning paradigms.

Fig.~\ref{fig:qualitative} presents qualitative comparisons across diverse environments, including social environments~(SCAND), indoor settings~(CEAR), off-road~(RELLIS-3D), and urban~(RUOS). First, regarding the LORT based configuration, the baseline consistently struggles to delineate boundaries across all environments.  

It often produces overly optimistic traversability maps, failing to suppress hazards. SyNeT+LORT significantly overcomes these limitations by leveraging synthetic negatives to enforce clearer separation between traversable and non-traversable regions, such as logs in off-road settings or cluttered objects in urban environments.

For the VS based configuration, the baseline already exhibits robust performance due to the semantic capabilities of the SAM encoder. However, in challenging scenarios like RUOS, characterized by diverse textures and visual ambiguity, SyNeT+VS further refines the decision boundaries. 

SyNeT is effective not only in improving weak baselines but also in correcting strong ones in complex real-world environments.

\begin{figure}[t]
    \centering
    \captionsetup{font=small}
    \includegraphics[width=0.48\textwidth]{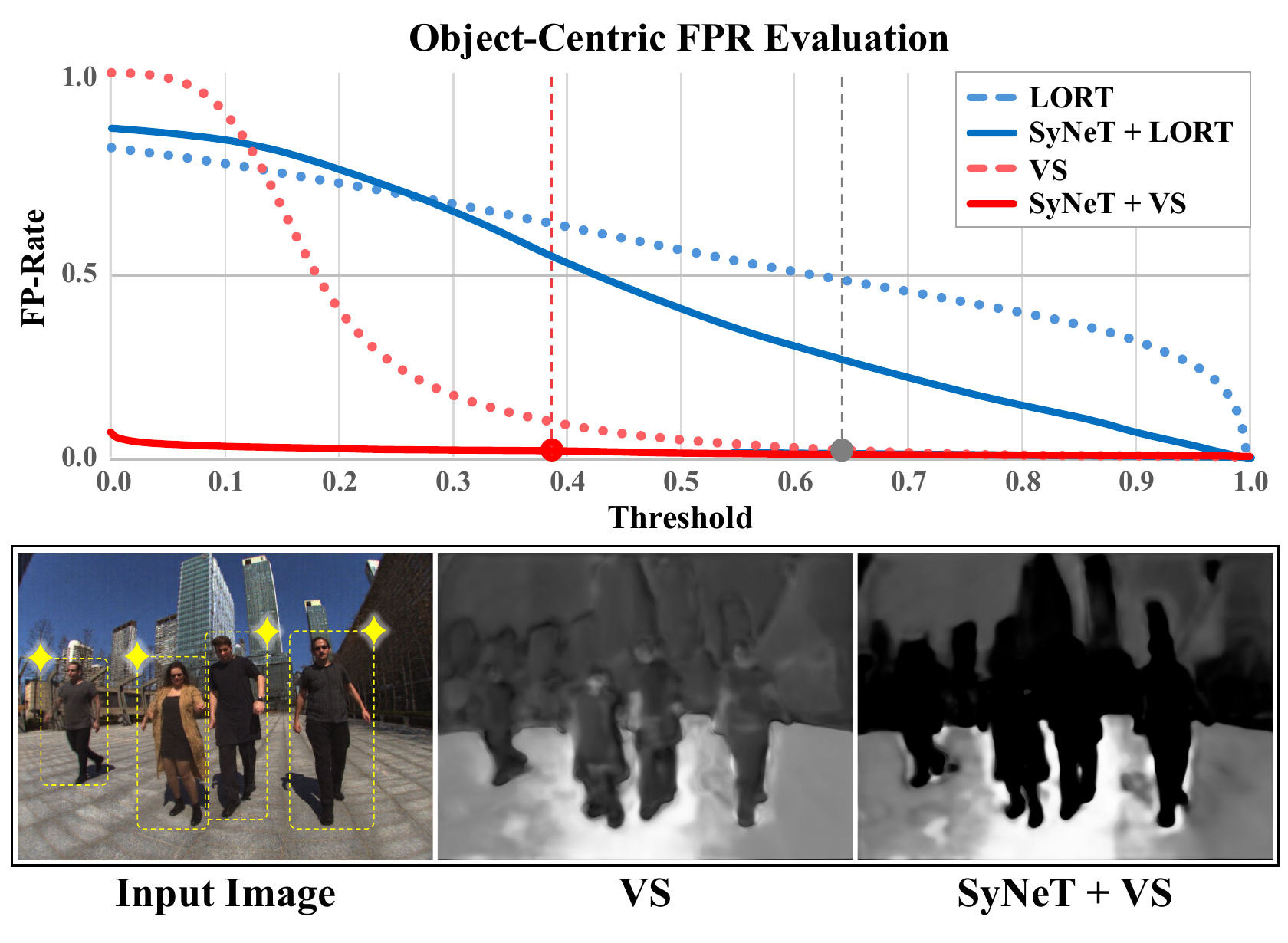}
    \caption{(Top) Object-centric FPR evaluation for all methods.
    (bottom) Qualitative results on RUOS convention center scene with synthetic negatives.}
    \vskip -1em
    \label{fig:obj-centric}
\end{figure}

\subsection{Object-Centric FPR Evaluation}

Conventional pixel-wise traversability evaluation relies heavily on manual annotation of non-traversable regions, which creates a significant bottleneck in terms of scalability and cost. To address this, we propose a novel annotation-free evaluation approach that leverages the automatically generated masks of synthetic negatives. Since these objects are explicitly synthesized as non-traversable, their masks serve as reliable pseudo-ground truth without requiring human labeling. Specifically, we compute the FPR over pixels within these synthesized regions and analyze the sensitivity of the model across varying decision thresholds. This evaluation directly measures how reliably each model identifies synthetic negatives as non-traversable, while incurring no additional labeling cost.

Fig.~\ref{fig:obj-centric} illustrates the object-centric evaluation using $400$ Synthetic dataset in RUOS, plotting the FPR against varying decision thresholds. The curves provide deeper insight into the performance metrics reported in Table~\ref{tab:rellis_main}. The baselines show relatively slow convergence, maintaining high error rates at lower thresholds, which indicates that achieving a reasonable FPR requires sacrificing traversability~(\textit{i.e.}, using an overly conservative threshold). In contrast, our SyNeT achieves an exceptionally low FPR almost immediately from the lowest thresholds. This stable curve explains the complementary quantitative results in Table \ref{tab:rellis_main}. It confirms that our SyNeT's ability to minimize false negatives is robust and insensitive to threshold variations, allowing for reliable traversability estimation.
\begin{figure}[t]
    \centering
    \captionsetup{font=small}
    \includegraphics[width=0.48\textwidth]{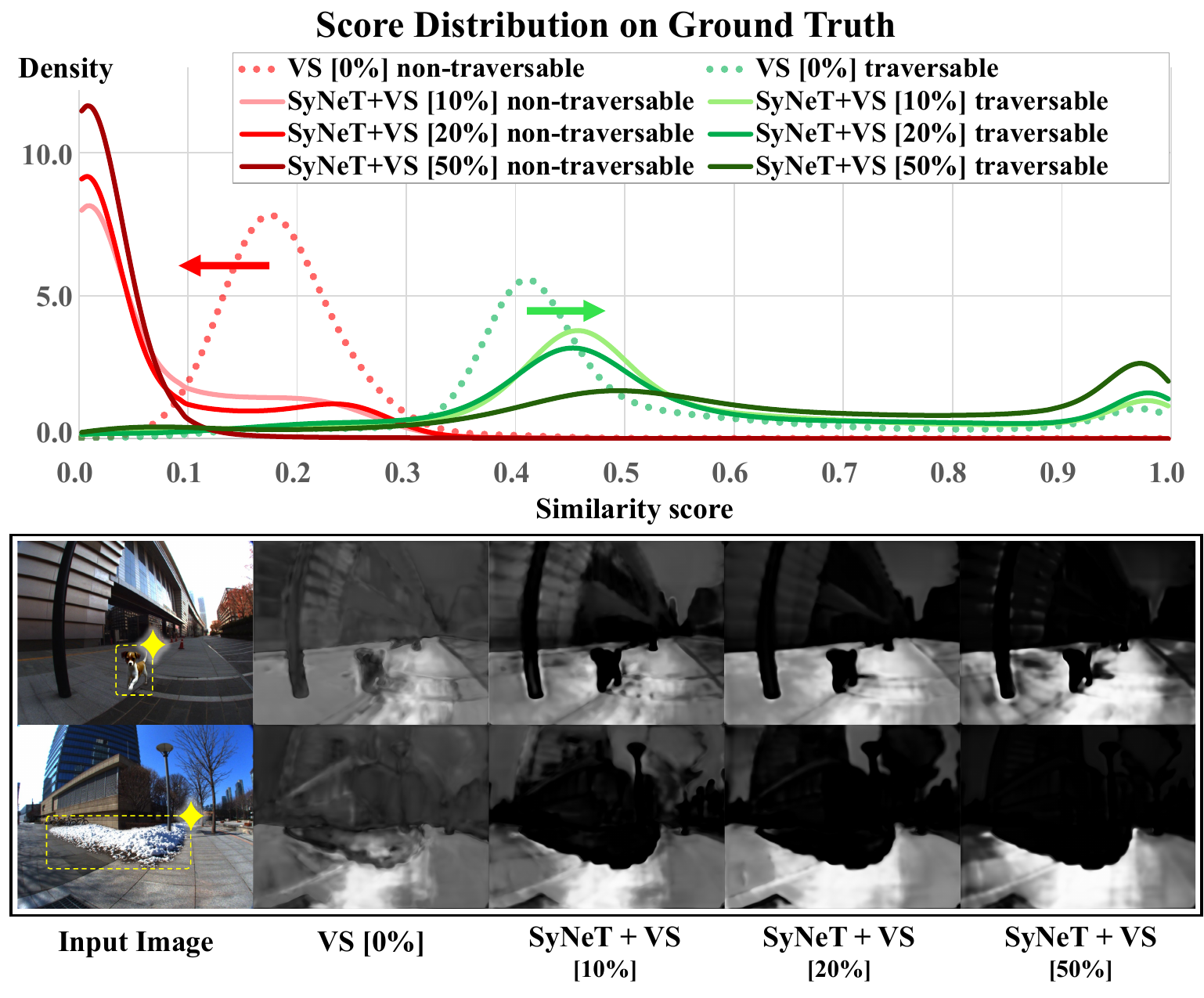}
    \caption{Effect of synthetic negatives ratio. (Top)~Estimated probability density functions of similarity scores computed over ground-truth traversable and non-traversable pixels. (Bottom)Traversability score maps. Compared to VS~(0\%), SyNeT+VS shifts the non-traversable distribution left and the traversable distribution right, increasing separation and reducing overlap.}
    \label{fig:ab1}
\end{figure}

\subsection{Ablation Studies}
\subsubsection{Effectiveness of Synthetic Negative Ratio}
\label{ablation1}
We vary the proportion of training images augmented with synthetic negatives~($0$\%, $10$\%, $20$\%, $50$\%) and observe a monotonic improvement in class separation. 
As the ratio increases, the non-traversable similarity distribution shifts left~(lower scores) while the traversable distribution shifts right~(higher scores), reducing overlap~(Fig.~\ref{fig:ab1}). 
This separation is reflected in fewer false positives around boundaries of non-traversable regions and cleaner suppression within these regions in the score maps. 
This progressive separation suggests that synthetic negatives help reduce the uncertainty of non-traversable regions by providing more explicit and consistent negative supervision.
In practice, $10\sim20$\% already yields most of the gain, while $50$\% shows diminishing returns and higher variance for the traversable class~(green), likely due to heavier augmentation pressure. 
We use $20$\% as the default in this experiment since it balances separation with training stability and preserves the appearance statistics of the base dataset.

\subsubsection{Effectiveness of Synthetic Negative Quality}
\label{ablation2}
\begin{figure}[t]
    \centering
    \captionsetup{font=small}
    \includegraphics[width=0.48\textwidth]{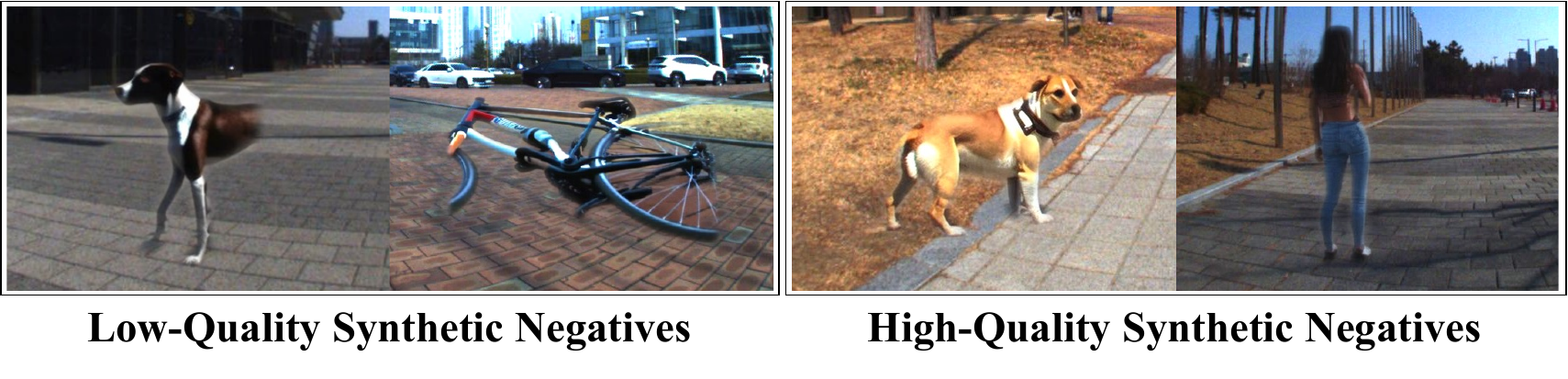}
    \caption{Examples of low-quality and high-quality synthetic negatives generated by the diffusion-based inpainting pipeline.}
    \label{fig:ab2}
\end{figure}

Because of limitations of the diffusion-based inpainting pipeline, some synthetic negatives exhibit low visual quality, such as cropped appearances or geometric distortions, as illustrated in Fig.~\ref{fig:ab2}. 
To analyze the impact of such artifacts, we compare two training settings with the same synthetic negative ratio and the same number of inserted negatives per class ($250$ each for bicycle, box, dog, and people). 
One model is trained using manually selected high-quality synthetic negatives, while the other uses all generated negatives without manual filtering.

As reported in Table~\ref{tab:ab_2}, both settings indicate that visual imperfections in the synthetic negatives have little impact on traversability performance under our learning framework.
We attribute this robustness to the fact that our model primarily learns pixel-level traversability cues rather than precise object geometry.

These results suggest that strict visual fidelity of synthetic negatives is not a critical requirement for effective negative supervision. 
Nevertheless, improvements in diffusion-based inpainting quality may further enhance training stability and visual realism, and we leave this as a direction for future work.

\begin{table}[t]
\centering
\scriptsize
\captionsetup{font=small}
\caption{Quantitative results under different synthetic negative quality settings on the RUOS}
\setlength{\tabcolsep}{8pt}
\begin{tabular}{l|c|c|c|@{}}\toprule
 Methods [Low-Quality / High-Quality]
 & AUROC$\uparrow$ & MaxF$\uparrow$  & AP$\uparrow$ \\
\midrule
VS
[ – / – ] & 0.958 & 0.949 & 0.983 \\
\textbf{SyNeT}+VS 
[$80$\% / $20$\%] & 0.982 & 0.960 & 0.993 \\
\textbf{SyNeT}+VS
[$0$\% / $100$\%] & 0.973 & 0.954 & 0.990 \\

\bottomrule
\end{tabular}
\vspace{-4mm}
\label{tab:ab_2}
\end{table}

\begin{table}[t]
\centering
\scriptsize
\caption{
On-robot performance comparisons of SyNeT.
}
\setlength{\tabcolsep}{3.8pt}
\begin{tabular}{l|c|c||l|c|c}
\toprule

Methods & MaxF$\uparrow$ & FPS$\uparrow$ &
Methods & MaxF$\uparrow$ & FPS$\uparrow$ \\
\midrule

\textbf{SyNeT}+LORT \textbf{(FP32)} & 0.951 & 29.732 &
\textbf{SyNeT}+VS \textbf{(FP32)}   & 0.945 & 2.264 \\

\textbf{SyNeT}+LORT \textbf{(INT8)} & 0.935 & 11.619 &
\textbf{SyNeT}+VS \textbf{(INT8)}   & 0.945 & 1.732 \\

\bottomrule
\end{tabular}
\label{tab:trt}
\end{table}

\subsection{On-robot Deployment}
To show the practical feasibility of traversability estimation methods on real robotic platforms, we benchmark our SyNeT on two representative systems: an RTX A6000 GPU (FP32) as a reference platform and an NVIDIA AGX Orin (INT8) as the actual on-robot embedded system. We convert the network models using quantization and TensorRT engines, and measure their inference speed and accuracy under on-robot constraints. Table~\ref{tab:trt} show the gap between offline GPU performance and embedded real-time execution, while demonstrating that TensorRT-based optimization enables practical on-robot inference without prohibitive accuracy degradation. We publicly release the TensorRT conversion and benchmarking code.

\section{CONCLUSIONS}
In this work, we proposed SyNeT, a training strategy that addresses the lack of explicit negative supervision in traversability learning.
SyNeT enables the model to learn more reliable representations of non-traversable regions, leading to clearer and more stable traversability estimation.
We validated the effectiveness of this approach through experiments on diverse public and self-collected datasets, covering indoor, off-road, and socially dynamic environments.
In addition, we introduced an object-centric FPR-based evaluation approach that provides an indirect yet practical way to analyze model behavior on synthetic negative regions without additional labeling.
As future work, we aim to reduce the need for manual tuning of synthetic negative design choices and to extend SyNeT to multi-modal settings, such as depth and infrared sensing, to further improve robustness under challenging perception conditions.




\bibliographystyle{IEEEtran}
\bibliography{mybib.bib}

\end{document}